\pdfoutput=1

\documentclass[11pt]{article}

\usepackage[]{ACL2023}

\usepackage{times}
\usepackage{latexsym}

\usepackage[T1]{fontenc}

\usepackage[utf8]{inputenc}

\usepackage{microtype}

\usepackage{inconsolata}

\usepackage{amsmath,bbm}
\usepackage{booktabs}
\usepackage{graphicx}
\graphicspath{{figures/}}
\DeclareGraphicsExtensions{.pdf,.png,.jpg}

\newcommand{\MNAME}{{\em{FiRo}}}
\title{\MNAME: Finite-context Indexing of Restricted Output Space\\ for NLP Models Facing Noisy Input}

\author{Minh Nguyen$^{1}$~\thanks{\;\;This work was initiated during internship at A$^*$STAR.}\qquad Nancy F. Chen$^{2,3,4}$\\
$^{1}$ Cornell University, USA \\
$^{2}$ Institute for Infocomm Research (I2R), A$^{*}$STAR, Singapore \\
$^{3}$ CNRS@CREATE, Singapore \\
$^{4}$ Centre for Frontier AI Research (CFAR), A$^{*}$STAR, Singapore
}

\begin{document}
\maketitle
\begin{abstract}
NLP models excel on tasks with clean inputs, but are less accurate with noisy inputs.
In particular, character-level noise such as human-written typos and adversarially-engineered realistic-looking misspellings often appears in text and can easily trip up NLP models.
Prior solutions to address character-level noise often alter the content of the inputs (low fidelity), thus inadvertently lowering model accuracy on clean inputs.
We proposed \MNAME, an approach to boost NLP model performance on noisy inputs without sacrificing performance on clean inputs.
\MNAME~sanitizes the input text while preserving its fidelity by inferring the noise-free form for each token in the input.
\MNAME~uses finite-context aggregation to obtain contextual embeddings which is then used to find the noise-free form within a restricted output space.
The output space is restricted to a small cluster of probable candidates in order to predict the noise-free tokens more accurately.
Although the clusters are small, \MNAME's effective vocabulary (union of all clusters) can be scaled up to better preserve the input content.
Experimental results show NLP models that use \MNAME~outperforming baselines on six classification tasks and one sequence labeling task at various degrees of noise~\footnote{Code is available at~\url{https://github.com/mnhng/FiRo}}.
\end{abstract}

\section{Introduction}\label{sec:intro}
Extensive use of pretrained language models~\cite{radford2018improving,devlin2019bert,liu2019roberta} has led to impressive performance on clean text.
However, these models are not robust to natural noise (e.g.\ irregular capitalization, misspellings, creative mix of characters and digits) and adversarial noise~\cite{pruthi2019combating}.
Thus, they often underperform when facing noisy inputs (e.g.\ social media text) during deployment~\cite{rosenthal2017semeval,belinkov2018synthetic}.
\begin{figure}[ht]
\centering
\includegraphics[width=\linewidth]{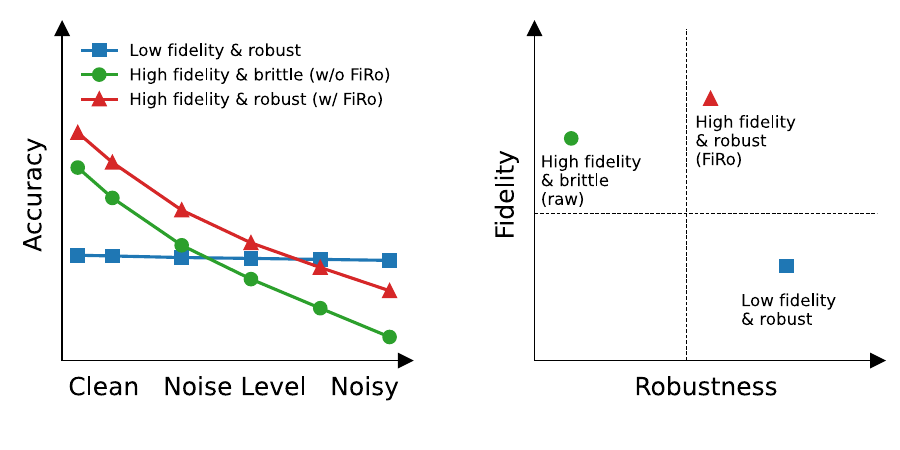}
\caption{\MNAME~has better better fidelity-robustness trade-off pre-processing text than other approaches do (including not doing pre-processing, i.e.~\emph{raw}).
Thus, NLP models when using \MNAME~pre-processed text do better on both clean and noisy inputs than when using text pre-processed by other approaches.
Text pre-processed by low-fidelity approaches may lead to poor performance on clean inputs while not doing pre-processing may lead to poor performance on noisy inputs.
Also see Section~\ref{ssec:rob_fid_res} for actual fidelity-robustness estimation.}\label{fig:intro_rob_fid}
\end{figure}

\begin{table*}[h]
\normalsize
\centering
\begin{tabular}{ll}
Original & Alex Trebek, host of Jeopardy!, is recovering from a minor heart attack in Los Angeles. \\
RobEn & alex \textbf{think}, \textbf{heart} of jeopardy!, is recovering from a \textbf{mother} heart attack in \textbf{less} angeles. \\
scRNN & \textbf{[U]} \textbf{[U]}, host of \textbf{[U]}!, is \textbf{[U]} from a minor heart attack in a los angeles. \\
\midrule
Original & \textbf{Tom} testified against John. \textbf{Tom} refused to turn on his friend. (\textbf{NLI label:} Contradiction) \\
Perturbed & \textbf{Tom} testified against John. \textbf{Tim} refused to turn on his friend. (\textbf{NLI label:} Neutral?)
\end{tabular}
\caption{Impacts of altering input. (Above) Information lost due to robustification.
RobEn replaces words with non-synonyms (e.g.\ host to \textbf{heart}).
scRNN replaces infrequent words with UNK (\textbf{[U]}) tokens.
(Below) Label flips due to adversarial noise altering a single word (invalid constant ground-truth label assumption).}\label{tbl:input_changes}
\end{table*}
Deployed models which analyze user inputs need to do well on both clean and noisy inputs.
Thus, models need to balance between the trade-offs of (1) sensitivity to semantic differences and (2) robustness to noise.
A model sensitive to even minor input changes can differentiate semantic changes but is also not very robust.
A model that always output the same prediction regardless of the inputs is extremely robust~\cite{jones2020robust} yet is of little use because it can only make trivial predictions.
Models can be robustified by training with an additional denoising objective (e.g.~BART~\cite{lewis2020bart}).
However, they may need to be retrained to cope with additional types of input noise (e.g.~noise faced when adapting to text input in new domains) and retraining could be costly because of the large number of parameters in these models.
In contrast, lightweight methods such as Spell correctors~\cite{pruthi2019combating} and Robust Encoding~\cite{jones2020robust} can be adapted quickly, cheaply, and independently of the NLP models to cope with additional types of input noise.

Spell correctors~\cite{pruthi2019combating} and Robust Encoding~\cite{jones2020robust} modify the inputs to remove variations due to noise (see Table~\ref{tbl:input_changes}).
Such modification may reduce semantic fidelity and make NLP models unable to perceive the semantic differences between text inputs.
For example, spell correctors often have limited vocabulary so they replace low-frequency and OOV words with UNK (unknown) tokens while Robust Encoding may map non-synonymous words to the same token.
This could lower the accuracy of downstream NLP models.
Figure~\ref{fig:intro_rob_fid} illustrates this fidelity-robustness trade-off.
Robust approaches (red and blue) lead to better downstream performance on noisy inputs while high-fidelity approaches (red and green) are more suitable than low-fidelity ones for clean inputs.
A high-fidelity and robust approach leads to good performance across the noise spectrum and would be ideal for deployment.

We propose \MNAME~(stands for Fidelity-Robustness), a fidelity-preserving neural pre-processor that helps NLP models cope with input character-level noise.
Given a noisy sequence of words, \MNAME~predicts the words' identities.
\MNAME~can help downstream models achieve high accuracy on both clean and noisy inputs.
Instead of using a common softmax covering all vocabularies, \MNAME's output space is input-specific and is restricted to only probable vocabularies.
The restricted output space makes \MNAME's output less susceptible to input noise.
Although the vocabulary size at each position is small, the effective model's vocabulary size (union of all softmaxes) can be sufficiently large.
Since \MNAME~can scale up the effective vocabulary size with minimal penalty on robust lexical prediction accuracy, \MNAME~covers more low-frequency and OOV words than prior models, thus better maintains input fidelity.
\MNAME~indexes into the restricted output spaces using contextual input token embeddings.
However, context window is finite instead of spanning the whole sequence so as to localize the effect of input noise.
Experiment results show that models that use \MNAME~achieve better results on six classification tasks and one sequence labeling task at various levels of character-level noise.

\section{Background}
\subsection{Realistic Imperceptible Character-Level Noise}
Adversarial noise can flip models' prediction while being \emph{imperceptible} to humans~\cite{szegedy2013intriguing}.
Since the noise is \emph{imperceptible} to humans, humans' prediction is invariant to the existence of the noise.
Thus, if models change their prediction as the result of the injected \emph{imperceptible} noise, they are not robust.
However, if adversarial noise was \emph{perceptible}, this type of robustness evaluation based on the invariant humans' prediction assumption might be invalid.
This is because humans' prediction could have changed as humans perceive the input difference (see Table~\ref{tbl:input_changes}).
In NLP, it is non-trivial to design \emph{imperceptible} adversarial noise~\cite{zhao2018generating}, since sentence-level noise~\cite{jia2017adversarial} or word-level noise~\cite{glockner2018breaking} are perceptible to humans~\cite{alzantot2018generating}.
In contrast, character-level noise~\cite{ebrahimi2018hotflip,belinkov2018synthetic} could be imperceptible to humans, as psycholinguistic studies demonstrated that humans may not be affected by jumbled internal characters~\cite{rawlinson1976significance,mccusker1981word}.

Yet, experts disagree about what level of character-level noise would qualify as \emph{perceptible}.
\citet{rawlinson1976significance} and \citet{perea2002does} suggested that humans are unaffected by character-level noise created by permuting internal characters, altering font size, or mixing cases (capitalization).
However, \citet{mayall1997disruption,davis2003psycholinguistic}, and \citet{rayner2006raeding} showed that mixing cases and character swaps would be \emph{perceptible} since it sometime causes humans to fail to comprehend the text.
When noise that may cause comprehension failure is injected into the text, it is unreasonable to expect the same prediction from both humans and models.
Given the lack of consensus, evaluation at one level of noise is inadequate since the chosen level may result in comprehension failure in humans, resulting in misleading conclusions.
We evaluated our models using multiple levels of character-level noise to analyze model performance when faced with noisy input.

\subsection{Controlled Noise Injection}
As it is difficult to gather human-written text at different noise levels, we conducted experiments using data injected with different levels of adversarial noise.
Adversarial noise is generated based on human error patterns (see Section~\ref{ssec:attack_param}) so it can appear naturalistic.
However, results at high noise level must be interpreted with caution as ground-truth labels may flip unknowingly (see Table~\ref{tbl:input_changes}).

Adversarial noise can be created using white-box attacks or black-box attacks.
In white-box attacks~\cite{goodfellow2014explaining}, attackers have access to either (a) the attacked model architecture and parameters, or (b) unlimited number of examples labeled by the attacked model.
In black-box attacks~\cite{papernot2017practical}, attackers have no access to the attacked model parameters and only a limited number of examples labeled by the attacked model.
Due to this constraint, an auxiliary model is usually needed to craft black-box attacks.
In this work, we consider both black-box and white-box attacks since they are complementary.
White-box attacks is harder to execute since the attackers must first gain access to the targeted ML model or collect a large labeled training set~\cite{papernot2017practical}.
Black-box attacks are easier to execute but may be less effective.
Nevertheless, commercial systems have been attacked successfully using only black-box attacks~\cite{liu2016delving}.

\subsection{Robustify Against Character-Level Noise}
Although adversarial training can theoretically robustify models against character-level noise~\cite{liu2020co,li2020context,zhao2021robust,si2020better}, in practice its impact can be limited~\cite{pruthi2019combating,jia2019certified} as old weaknesses can resurface during training.
An alternative is to integrate inductive biases such as character-permutation invariant representation~\cite{belinkov2018synthetic,wang2020learning,liu2020joint,sankar2021device} into models.
For example, RoVe~\cite{malykh2018named,malykh2019robust,malykh2023robust} generates word embeddings that are invariant to character swaps by encoding each word as a bag of characters.
Another example is Robust Encoding~\cite{jones2020robust}, a representation that is invariant to most perturbations within one-character edit distance.
Nevertheless, both lines of work require re-training the models~\cite{eshel2017named,michel2018mtnt,ribeiro2018semantically} which may be inconvenient or costly, especially as NLP models grow rapidly in size.
General plug-and-play robustification techniques~\cite{contractor2010unsupervised} are more appealing since they can be deployed right away, regardless of the tasks or the models.
\citet{pruthi2019combating} proposed a plug-and-play model to sanitize the input text, obviating the need for re-training downstream models.
However, this model struggles with unseen words due to its limited vocabulary.
While~\citet{pruthi2019combating} did propose back-off strategies to handle unseen words, such strategies may work only in specific tasks and may compromise fidelity.

\begin{figure}[ht]
\centering
\includegraphics[width=\columnwidth]{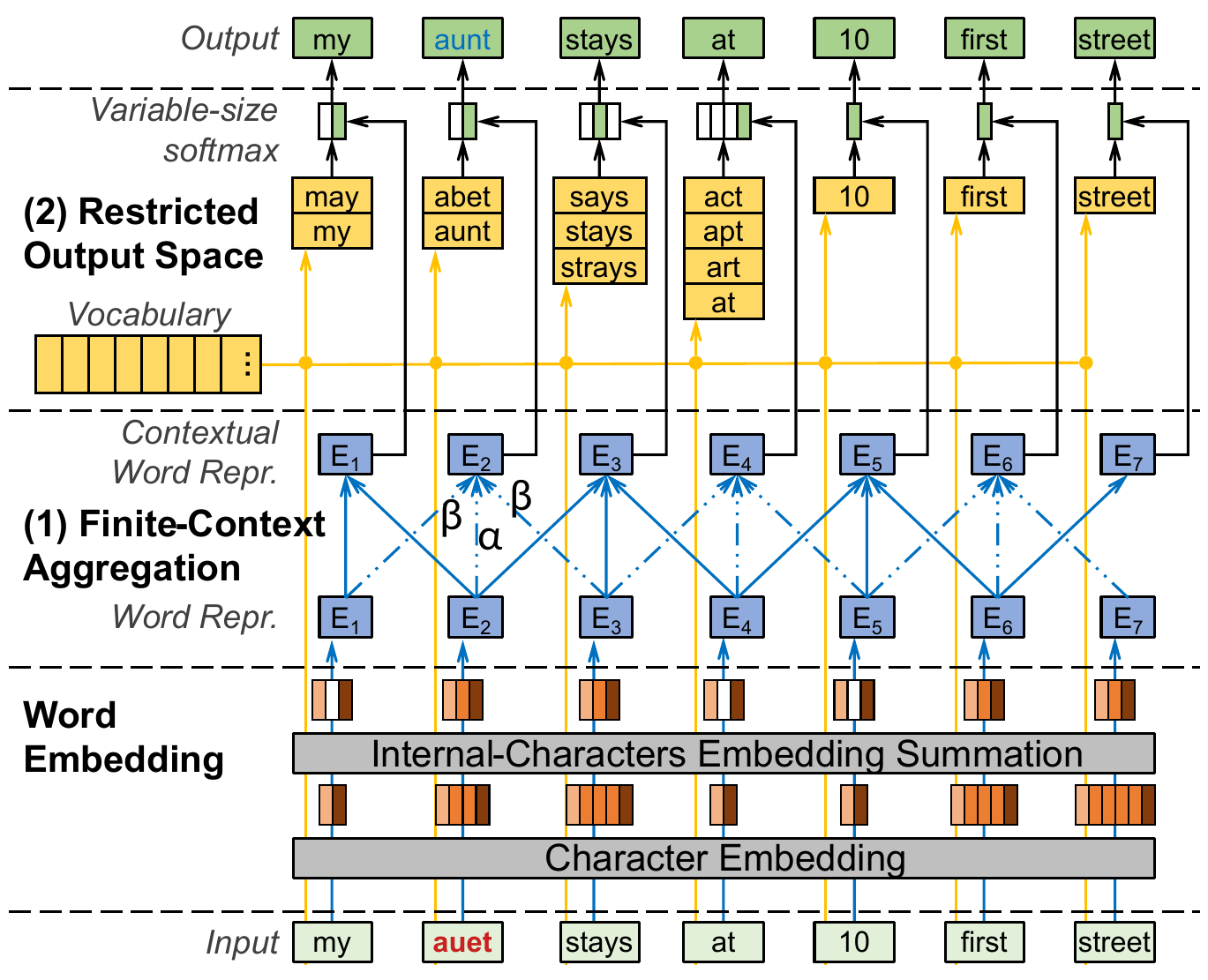}
\caption{\MNAME.
(1) Construct contextual embedding by finite-context aggregation.
(2) Retrieve cluster of words similar to input words (restricted output spaces). Predict output word from cluster using contextual input embedding.
The cluster sizes (<100) are much smaller than the vocabulary size (100k).}\label{fig:model}
\end{figure}
\section{Method}\label{sec:method}
\subsection{Proposed Approach}\label{ssec:model}
Figure~\ref{fig:model} outlines \MNAME~which processes noisy inputs in two steps: (1) Finite-Context Aggregation and (2) Restricted Output Space Indexing.
Although \MNAME~operates at the word-level internally, \MNAME~outputs text sequences (by concatenating the output words) that can be analyzed by both word-based or subword-based NLP models.

\paragraph{Finite-Context Aggregation:}
First, the input string is tokenized into words and the words are turned into embeddings using the same approach used by \citet{sakaguchi2017robsut}.
Specifically, all characters in a word are mapped into character embeddings.
Subsequently, the first character's embedding, the last character's embedding, and the average of the internal characters' embeddings are concatenated to form the word's embedding.
Unlike word embeddings constructed using CNN or RNN, these word embeddings are invariant to noise induced by letter swaps~\cite{sakaguchi2017robsut,belinkov2018synthetic}.
Contextual input embeddings are then weighted averages of adjacent word embeddings (finite-context).
Let $h_i$ be the word embedding at position $i$, the contextual embedding at position $i$ is defined as $\alpha h_i + 0.5(1-\alpha)(h_{i-1} + h_{i+1})$.
The coefficient $\alpha$ is learned during \MNAME~training.

In humans, accurate word recognition also requires identifying constituent characters and surrounding context~\cite{whitney2004does}.
Higher-order linguistic knowledge and lexical context can refine the representation of individual characters in words and correct for perturbations induced by noise~\cite{heilbron2020word}.
While global self-attention~\cite{vaswani2017attention} is often used for contextual embeddings (a word attending to all other words), it will allow perturbation from a single position to potentially spread to all positions, leading to low robustness.
Even though local self-attention~\cite{yang2018modeling} is more robust than global self-attention as the effect of noise is curtailed to only local words, perturbations may still result in noisy keys that cause self-attention to fail to aggregate information from neighbors.
By using local weighted averages, finite-context aggregation localizes the impact of perturbation while ensuring that contextual information from neighboring words is always considered (also see Section~\ref{ssec:finite_context}).

\paragraph{Restricted Output Space Indexing:}
At each position, the output space is restricted to a small variable-size cluster containing words similar to the input word.
Like~\citet{jones2020robust}, similarity is defined as one edit distance apart.
In particular, for an input word A, all words that are within one edit distance of A are put into the cluster.
Indexing into the output space is done by taking the softmax of the dot product of the input word's contextual embedding and the embeddings of words in the cluster.
\MNAME~then outputs the word in the cluster with the highest probability.
As the cluster sizes vary, so do the softmax sizes.
As the clusters are much smaller in size than the full vocabulary (<100 vs 100k), perturbing input words leads to limited and more predictable change to \MNAME's output.
Despite using small clusters, \MNAME's effective vocabulary (union of all clusters) is considerable and can be scaled up (with minimal increase in cluster sizes) to avoid predicting UNK for infrequent and OOV words.
Thus, \MNAME~can preserve input fidelity better while being robust to perturbations.
In contrast, while Robust Encoding also use clusters to map input words to output words, its mapping ignores context, leading to input fidelity loss.

\begin{figure*}[ht]
\centering
\includegraphics[width=\linewidth]{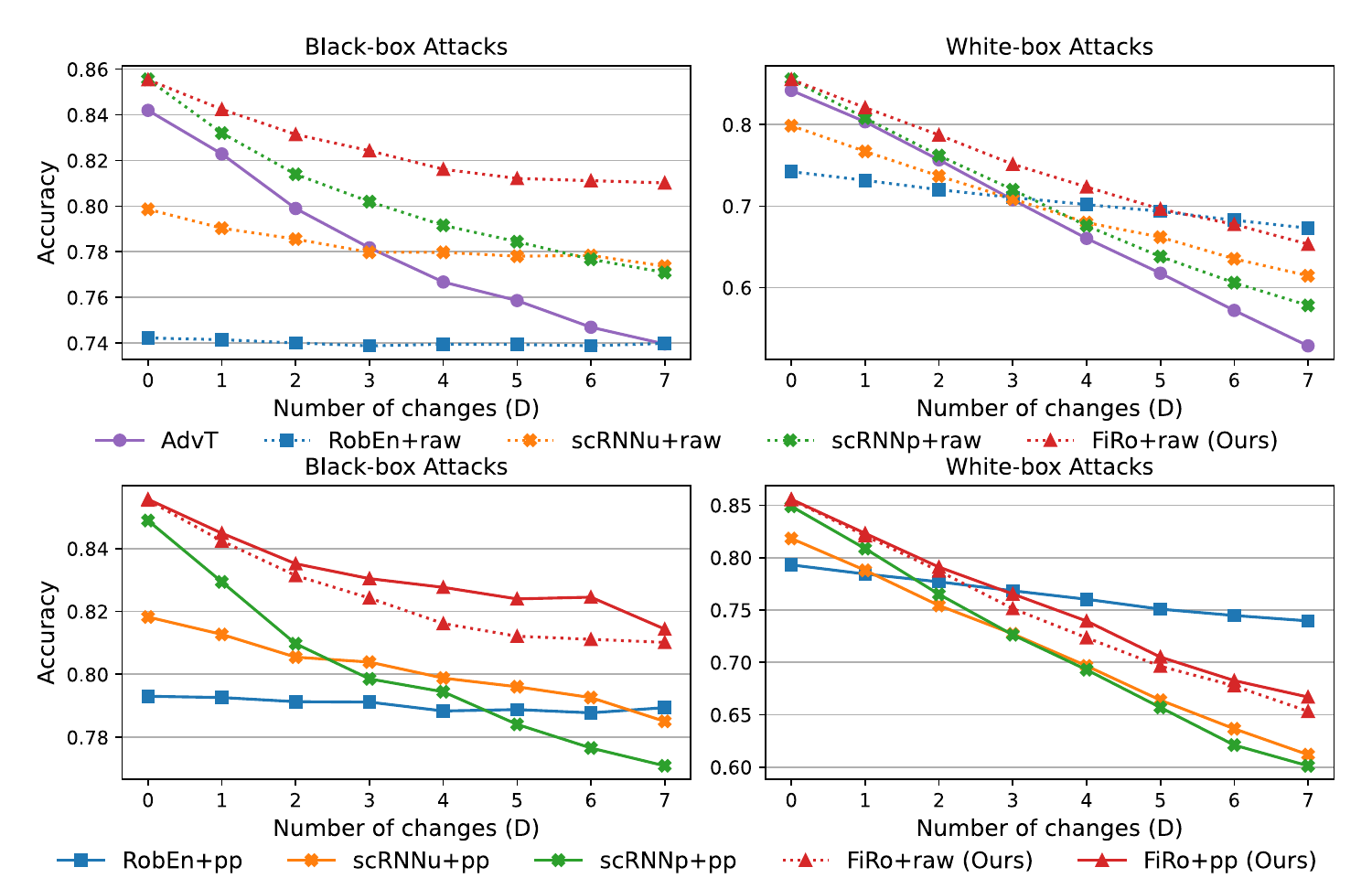}
\caption{Average GLUE accuracy as the number of adversarial changes introduced into the input text varies.
``+raw'': BERT fine-tuned using raw text.
``+pp'': BERT fine-tune using \textbf{p}re-\textbf{p}rocessed text.
Best approach from top panels are included in bottom panels for comparison.
Using \MNAME~leads to higher accuracy than baselines in general.}\label{fig:main_result}
\end{figure*}

\subsection{Baselines}\label{ssec:baselines}
We compared \MNAME~against 4 baselines: adversarial training, two variants of the \emph{scRNN} spell correctors~\cite{pruthi2019combating}, and a variant of Robust Encoding called Agglomerative Cluster Encodings~\cite{jones2020robust}.
\begin{itemize}
\itemsep0em
    \item \emph{AdvT}: Adversarial training
    \item \emph{scRNNu}: \emph{scRNN}, predict UNK for OOV
    \item \emph{scRNNp}: \emph{scRNN}, let OOV pass through
    \item \emph{RobEn}: Robust Encoding
\end{itemize}
Adversarial training is an end-to-end approach that fine-tunes the NLP model to make it more accurate when facing noisy inputs.
In contrast, \MNAME~and the spell correctors do not change the NLP model's weights.
Between the spell correctors, \emph{scRNNp} is less robust since it lets OOVs pass through unmodified, exposing downstream models to (adversarial) noise.
However, \emph{scRNNu} is has lower fidelity since it maps OOVs to UNK.
Among the baselines, \emph{RobEn} is the most robust but also has the lowest fidelity (also see Section~\ref{ssec:rob_fid_res}).
Due to \emph{RobEn}'s low fidelity, a downstream model would not work out-of-the-box on text encoded using \emph{RobEn}.
Hence, for \emph{RobEn} specifically, downstream models are fine-tuned using \emph{RobEn} encoded text instead of original text (similar to \citet{jones2020robust}).
Although RoVe~\cite{malykh2023robust} and \MNAME~share some architectural similarities~\cite{sakaguchi2017robsut}, using RoVe to robustify pretrained language models is much harder.
This is because RoVe's outputs are sequences of embeddings instead of sequences of words.
Thus, combining RoVe with pretrained language models requires replacing the input embeddings of the pretrained language models which may lead to low performances for tasks with limited data for finetuning.
Given this limitation, RoVe was not chosen as a baseline.

\subsection{Implementation Details}\label{ssec:implementation}
We used PyTorch~\cite{paszke2019pytorch} and \emph{transformers}~\cite{wolf2020transformers} libraries in this work.
BERT is the NLP model and is fine-tuned for 3 epochs using a learning rate of $2e^{-5}$ and a batch size of 8 with AdamW~\cite{loshchilov2018decoupled}.
\MNAME~and scRNN are trained using Adam~\cite{kingma2015adam} using a batch size of 50 until convergence (about 10 hours using an NVIDIA TitanXp GPU).
We used the GLUE's training sets as data for training \MNAME~and scRNN.

The architecture of scRNN follows that in prior studies~\cite{pruthi2019combating,jones2020robust} with the vocabulary size set at 10,000.
scRNN is based on a bidirectional LSTM~\cite{graves2005framewise} with one layer of size 50.
Similar to~\citet{jones2020robust}, the 100,000 most frequent words from the COCA corpus~\cite{davies2008COCA} are used as RobEn's vocabulary.
\MNAME's vocabulary is the same as RobEn's.
For the adversarial training baseline, the fine-tuned BERT model is further trained using a equal mixed of normal and adversarial examples until there is no further increase in accuracy for a hold-out adversarial set of data.

\subsection{Parameters of Character-Level Attacks}\label{ssec:attack_param}
Assume that the attacked model is represented by a function $f$.
For an input $x$ with ground truth label $y$, the model predicts $f(x)$ as the label.
An adversarial input $x^*$ is an instance close to $x$, such that $x^*$ has the same ground truth as $x$, while $f(x^*) \neq y$ (\citet{szegedy2013intriguing,liu2016delving}).
The `closeness' between $x$ and $x^*$ is denoted as $d(x, x^*)$, the number of words that differ between $x$ and $x^*$ due to some character perturbations.
Let $A$ denote the search for an adversarial example, Equation~\ref{eq:wbox} and~\ref{eq:bbox} show the two types of attacks.
The auxiliary model used by the black-box attack is denoted $f^{aux}$.
Both attacks need to satisfy the `closeness' constraint in Equation~\ref{eq:constraint}.
The level of adversarial noise can be controlled by choosing the choice of character perturbations and the value of $D$ (number of allowable modified words).
\begin{align}
x_{white-box}^* = A(f, x, y) = arg_{x'} f(x') \neq y \label{eq:wbox}\\
x_{black-box}^* = A(x, y) = arg_{x'} f^{aux}(x') \neq y \label{eq:bbox}\\
d(x, x^*) \leq D \label{eq:constraint}
\end{align}

Adversarial attacks are valid when $x$ and $x^*$ have the same ground-truth label~\cite{szegedy2013intriguing,liu2016delving}.
For tasks like natural language inference (NLI), even a single letter substitution could violate the label-invariant assumption (see Table~\ref{tbl:input_changes}).
Thus, the types of character perturbations and the constant $D$ must be chosen carefully to avoid drawing invalid conclusions.

We followed~\citet{gao2018black,pruthi2019combating,jones2020robust} by crafting character-level attacks using four basic operations: (1) character substitution, (2) character deletion, (3) character insertion, and (4) swapping of two adjacent characters.
The operations must not cross word-boundary and the characters are picked at random.
The assumption that $x$ and $x^*$ have the same ground-truth label is more likely to be violated for larger $D$, so we evaluated for $D$ in the range from 0 to 7.

\subsection{Quantifying Robustness and Fidelity}\label{ssec:rob_fid}
We can empirically estimate (1) $\mathsf{Robustness}$ and (2) $\mathsf{Fidelity}$ of \MNAME~by comparing its denoised outputs against the clean input.
For each clean input $x$, a set of perturbed inputs $X^* = \{x^*\}$ is generated.
For each noisy input $x^*$ in this set, \MNAME~outputs a denoised version $z$.
This results in a set of denoised outputs $Z = \{z\}$ for each $x$.
The identity (i.e.\ $x$) is also included in $Z$.
$\mathsf{Robustness}$ quantifies how similar the denoised outputs are (Equation~\ref{eq:rob}), while $\mathsf{Fidelity}$ quantifies how closely denoised outputs match the clean input (Equation~\ref{eq:fid}).
Let $|Z|$ be the size of $Z$; $L$ be the length of $z$; $\mathsf{uniq}(Z)$ be the set of unique elements in $Z$; $\mathbbm{1}$ be the indicator function; and $z_i$ be the $i$th token in $z$.
\begin{align}
    \mathsf{Robustness} &= \frac{|Z| + 1 - |\mathsf{uniq}(Z)|}{|Z|} \label{eq:rob}\\
    \mathsf{Fidelity} &= \frac{1}{|Z|} \sum_{z\in Z} \frac{1}{L}\sum_{1\leq i\leq L} \mathbbm{1}_{\{z_i = x_i\}} \label{eq:fid}
\end{align}
$\mathsf{Robustness}$ and $\mathsf{Fidelity}$ range from 0 to 1.
$\mathsf{Robustness}$ is maximized when all elements in $Z$ are identical.
$\mathsf{Fidelity}$ is maximized when all elements in $Z$ are the same as $x$.
For multiple $x$, the $\mathsf{Robustness}$ and $\mathsf{Fidelity}$ values are averaged.
The same estimate can be calculated for the baselines.

Specifically, we estimated the empirical robustness and fidelity using the GLUE data.
For each input $x$, 10 noisy copies $x^*$ are created sequentially by sampling 10 positions in $x$ and inject character-level noise into the tokens at these positions.
Thus, the inputs $x^*$ have increasing level of noise.

\section{Experiments}
\subsection{GLUE Experiment Setup}\label{ssec:glue_setup}
Followed \citet{jones2020robust}, we experimented on six GLUE~\cite{wang2018glue} tasks: MRPC, MNLI, QNLI, QQP, RTE, and SST-2.
Approaches are evaluated using average task accuracy.
We used the BERT~\cite{devlin2019bert} base uncased model as the backbone for all six classification tasks.

For each task, BERT is fine-tuned using the training set and evaluated on the validation set.
For evaluation, the input text is first processed by \MNAME, scRNN, or RobEn before being passed to BERT.
There are two ways to fine-tune BERT.
The first is to fine-tune BERT using the raw GLUE text.
Thus, BERT is oblivious to the text pre-processor (i.e. \MNAME, scRNN, or RobEn) and fine-tuning does not have to be redone every time the text pre-processor is changed or improved.
The second is to fine-tune BERT using the pre-processed text (output of \MNAME, scRNN, or RobEn).
This allows BERT to adapt to the idiosyncrasies of the text pre-processor, resulting in more robust models although at the expense of frequently redoing fine-tuning.
We evaluated the approaches using both ways of fine-tuning.

Adversarial examples are found using beam-search with a beam size of 5 similar to~\citet{jones2020robust}.
For black-box attacks, beam search uses the backbone BERT model.
For white-box attacks, beam search uses the combined model comprising of the backbone and a defender (e.g. \MNAME~or RobEn).
The latter are white-box attacks because the combined model is queried without any limit (scaling linearly with the number of test examples).

\begin{figure*}[ht]
\centering
\includegraphics[width=\linewidth]{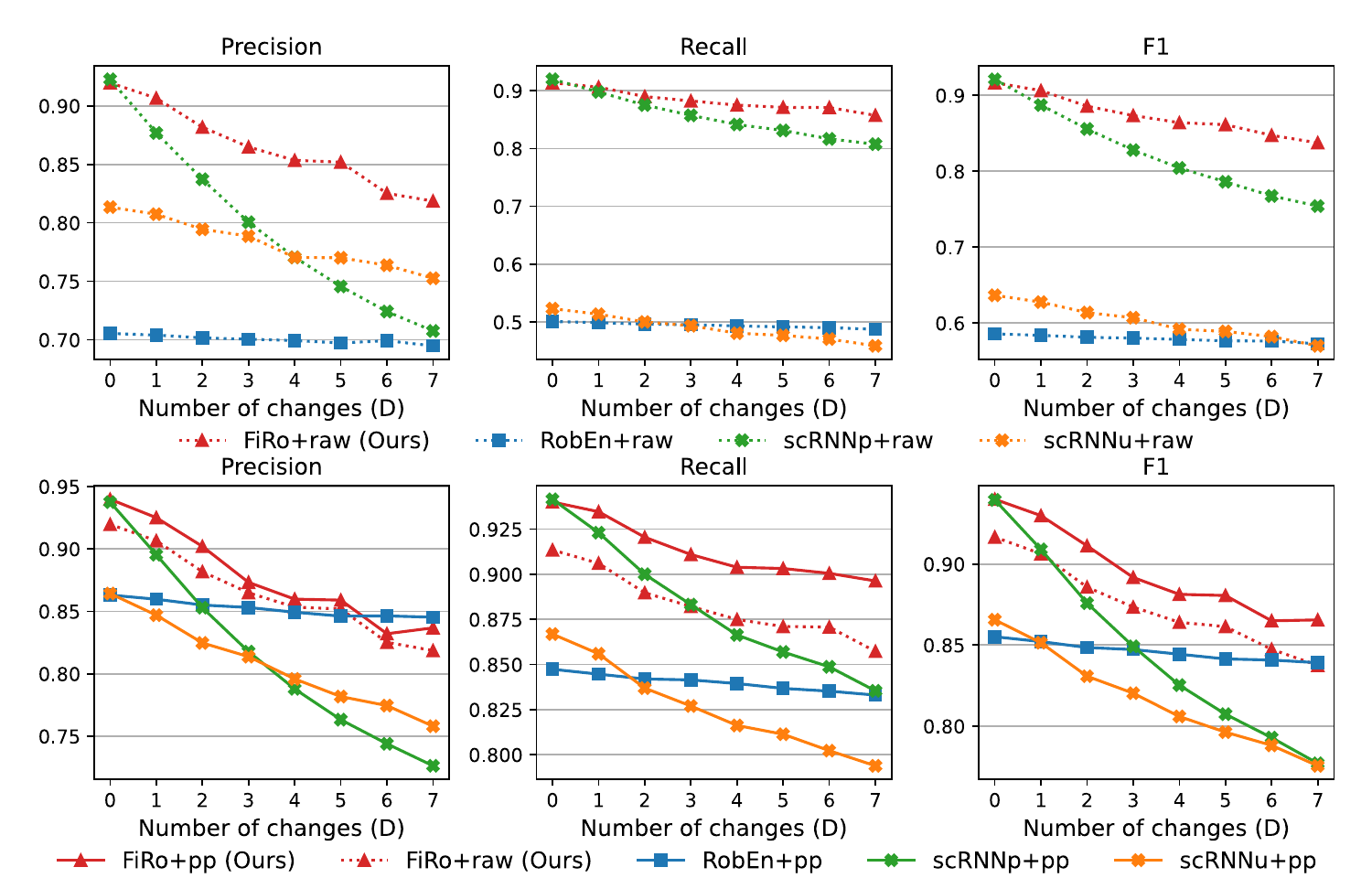}
\caption{CoNLL-2003 NER performance as a function of $D$.
``+raw'': BERT fine-tuned using raw text.
``+pp'': BERT fine-tune using \textbf{p}re-\textbf{p}rocessed text.
BERT using \MNAME~pre-processed text obtains higher F1 and recall.}\label{fig:ner_result}
\end{figure*}

\subsection{Results from GLUE Experiment}\label{ssec:main_result}
Figure~\ref{fig:main_result} shows the average accuracy of 6 GLUE tasks under different adversarial attacks.
Left panels show performance under black-box attacks.
Right panels show performance under white-box attacks.
BERT is fine-tuned using the raw GLUE text in the top panels while it is fine-tuned using the pre-processed text in the bottom panels.

For brevity, in this section, \MNAME~refers to the BERT model using \MNAME~preprocessed text and so on.
\MNAME~outperforms scRNNp, scRNNu and adversarial training baselines in all scenarios.
\MNAME~is better than RobEn when under black-box attacks.
When under white-box attacks, \MNAME~is better when the noise is low ($D\leq 5$ top right panel; $D\leq 3$ bottom right panel), but RobEn is better when the noise is high.
\MNAME~and scRNNp preserve the fidelity of the input well since they did well on clean text ($D=0$).
However, as scRNNp lets unrecognized words (e.g. OOVs or words modified by adversarial attacks) pass through unmodified, it is less robust.
In contrast, by supporting a large vocabulary using the scope output layer, \MNAME~can cover more infrequent words while also being quite robust.
\MNAME~and RobEn are more robust than the other approaches since their accuracy drops less rapidly as the noise level increases.

Another advantage of \MNAME~is that it can be used out-of-the-box.
When BERT is not adapted to the pre-processor models (top panels), \MNAME~is generally better than baselines.
In addition, the bottom panels show that \MNAME's performance when BERT is fine-tuned using raw GLUE text (dotted red line) is quite close to \MNAME's performance when BERT is instead of processed GLUE text (solid red line).
Being able to use \MNAME~straight away without having to redo fine-tuning the BERT model could lower the cost of NLP model deployment.

\subsection{NER Experiment Setup}
We use the CoNLL-2003 named entity recognition (NER) dataset~\cite{sang2003introduction} for this experiment.
We compare \MNAME~against scRNNu, scRNNp, and RobEn.
We also use the BERT~\cite{devlin2019bert} base uncased model as the backbone and reuse the models (\MNAME, scRNNu, and scRNNp) trained using the GLUE data.
Training of the backbone model is done similar to the procedure in the sequence classification experiment.
For sequence tagging, we only explore black-box attacks since, as far as we know, there is no prior work on conducting adversarial attack for tasks that are not classification.
Similar to the GLUE experiment, adversarial examples are found using beam-search with a beam size of 5.
For sequence tagging, the beam search's objective is maximizing the non-overlapped named entities between the ground truth entity set and the predicted entity set.

\subsection{Results from NER Experiment}\label{ssec:ner_result}
Figure~\ref{fig:ner_result} shows the performance under varying degrees of black-box attack ($D=0, 1, 2,\dots 7$).
\MNAME~obtains higher F1 score and higher recall than the baselines across all scenarios due to \MNAME's high fidelity.
In contrast, RobEn and scRNNu underperform in this task because they fail to preserve the fidelity of the inputs.
Although scRNNp does well for clean inputs ($D=0$), its performance degrades quickly as the noise level increases.

\begin{figure}[ht]
\centering
\includegraphics[width=\linewidth]{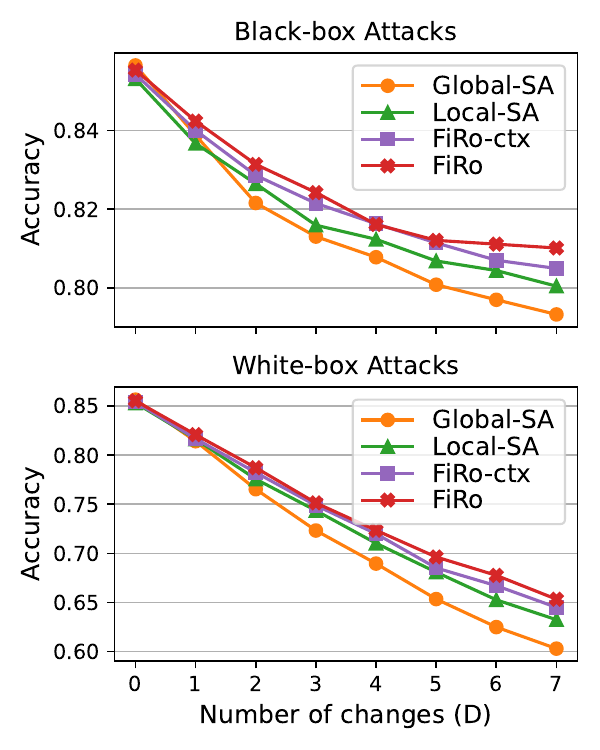}
\caption{Average GLUE accuracy.
Using finite-context aggregation (\MNAME) results in higher accuracy than using local or global self-attention (SA).}\label{fig:sa_ablate}
\end{figure}

\subsection{Spell Correction Experiment}
In the previous two experiments, \MNAME~was evaluated on data with synthetic character-level noise.
Thus, we additionally evaluate \MNAME's performance on a spell correction task with realistic character-level noise.
However, it must be emphasized that although \MNAME~functions like a spell corrector with the current cluster design (clusters are defined based on textual edit distance; Section~\ref{ssec:model}), \MNAME~is more general than a spell corrector because the cluster design can be adapted for other tasks (e.g.~spoken language processing tasks).

For this experiment, we use the GitHub Typo Corpus~\cite{hagiwara2020github} which comprises of typos (character-level noise) and grammatical errors collected from public code repositories on GitHub.
Since this corpus includes multi-lingual text and diverse types of noise (e.g.~character-level, word-level, capitalization noise), we exclude samples that are not typos in English.
Specifically, text sequences that are kept must be: (1) written in English, (2) with high probability of being typos, (3) with low perplexity ($\leq5$).
Consequently, the remaining 53,154 samples are used to test spell correction performance.
The models were trained using the GLUE data (see Section~\ref{ssec:implementation}).
\begin{table}
    \normalsize
    \centering
    \aboverulesep=0ex
    \belowrulesep=0ex
    \begin{tabular}{lccc}
        Method & Precision & Recall & F1 \\
        \midrule
        scRNNu	& 0.076 & 0.294 & 0.120 \\
        scRNNp 	& 0.306 & 0.298 & 0.302 \\
        \MNAME 	& \textbf{0.514} & \textbf{0.463} & \textbf{0.487}
    \end{tabular}
    \caption{Spell correction performance evaluated using word-level metrics (precision, recall, F1)}\label{tbl:spell_correction}
\end{table}

\subsection{Results from Spell Correction Experiment}\label{ssec:spell_correction}
Table~\ref{tbl:spell_correction} shows the spell correction performance of \MNAME~and scRNN.
RobEn is excluded because it is not devised as a spell corrector.
Although these models were trained using the GLUE data injected with synthetic character-level noise, they managed to achieve decent spell correction performance on the GitHub corpus which contained typos made by humans.
\MNAME~obtains higher precision, recall, and F1 score than scRNN.
The performance of \MNAME~could be improved further with better cluster design since the clusters used are based on textual similarity of one edit distance apart but the GitHub corpus definitely contains more diverse typos.

\section{Ablation}
\subsection{Finite-context Aggregation}\label{ssec:finite_context}
Figure~\ref{fig:sa_ablate} shows that without context aggregation (\emph{\MNAME-ctx}), \MNAME~performs worse.
Using global self-attention (\emph{Global-SA}) or local self-attention with the same neighborhood size as the finite-context aggregation (\emph{Local-SA}) results in less robust models.
Thus, finite-context aggregation seems to give the best overall performance across noise spectrum.

\begin{table}[ht]
\normalsize
\centering
\aboverulesep=0ex
\belowrulesep=0ex
\begin{tabular}{l|@{\hspace{.5em}}c@{\hspace{.5em}}c|@{\hspace{.5em}}c@{\hspace{.5em}}c@{\hspace{.5em}}c}
Method & Fi & Ro & Arith & Geo & Har \\
\midrule
RobEn	& 0.678 & 0.931 & 0.794 & 0.804 & 0.784 \\
scRNNu	& 0.863 & 0.811 & 0.837 & 0.837 & 0.837 \\
scRNNp 	& \textbf{0.938} & 0.684 & 0.801 & 0.811 & 0.791 \\
\MNAME 	& 0.910 & \textbf{0.962} & \textbf{0.936} & \textbf{0.936} & \textbf{0.935}
\end{tabular}
\caption{Estimates of $\mathsf{Robustness}$ (\emph{Ro}) and $\mathsf{Fidelity}$ (\emph{Fi}).
\emph{Arith}, \emph{Geo}, and \emph{Har} are the arithmetic, geometric, and harmonic means of $\mathsf{Robustness}$ and $\mathsf{Fidelity}$.}\label{tbl:rob_fid}
\end{table}
\subsection{Empirical Robustness-Fidelity Estimation}\label{ssec:rob_fid_res}
Table~\ref{tbl:rob_fid} reports the empirical $\mathsf{Robustness}$ and $\mathsf{Fidelity}$ and their arithmetic, geometric, and harmonic mean (also see Section~\ref{ssec:rob_fid}).
\MNAME~has the best $\mathsf{Robustness}$-$\mathsf{Fidelity}$ trade-off (highest arithmetic, geometric, and harmonic means).
\MNAME's $\mathsf{Fidelity}$ leads to good performance for clean input ($D=0$) in Section~\ref{ssec:main_result} and~\ref{ssec:ner_result}.
One limitation of the $\mathsf{Robustness}$ measure is that it cannot distinguish between ``trivially'' robust models and those that are genuinely robust.
A ``trivially'' robust but vacuous model can achieve very high $\mathsf{Robustness}$ by predicting the same output regardless of inputs.
Thus, $\mathsf{Robustness}$ should be considered in tandem with $\mathsf{Fidelity}$ instead of considered as a standalone metric.
For models with similar $\mathsf{Fidelity}$, those with higher $\mathsf{Robustness}$ would perform better overall and under noisy condition in particular.
For example, \MNAME~and scRNNp have similar $\mathsf{Fidelity}$ but since \MNAME~has higher $\mathsf{Robustness}$, \MNAME~beats scRNNp as the noise level increases (see Figure~\ref{fig:main_result}).

\section{Discussion}\label{sec:discussion}
In the tasks used to evaluate robustness, we only used synthetic noise.
However, the design of the synthetic noise is motivated by observations from naturally generated data, i.e.~noisy text written by humans.
Besides, while it is true that the noise characteristics do not change between training and test, all methods evaluated exploit this understanding about the noise characteristics in their model architecture/algorithm to enable more robust text processing.
Thus, \MNAME~does not benefit from any unfair advantages in this comparison setup.
Furthermore, Section~\ref{ssec:spell_correction} shows that \MNAME~trained on synthetic noise can generalize to human typos.

Robustification methods should avoid information loss (preserve fidelity) so as to be applicable to many different tasks.
Some tasks can be completed using only a few clues from the input, therefore loss of information does not affect task accuracy.
However, information loss can greatly affect performance in tasks that require exact phrasing (e.g.\ summarization, translation).
Since \MNAME~can preserve input fidelity better than other approaches, it can be applied to a wider variety of tasks.

\MNAME~achieved a reasonable level of robustness without sacrificing word recognition performance chiefly due to its restricted output space.
This allows \MNAME~to scale up the vocabulary size while still being robust to misspellings.
Extending \MNAME~to natural noise would require expanding the output space to cover phenomena such as abbreviations and word play using phonetic spelling (e.g.~using `b4' for `before', `gr8' for `great').
Orthographic and phonological similarity constraints have been explored to improve accuracy of correcting character-level misspellings in Chinese~\cite{nguyen2021domain}.
However, this study is only feasible because of the relatively clear orthographic and phonological relationships between Chinese characters~\cite{nguyen2018multimodal,nguyen2019hierarchical}.
Such endeavors for English is beyond the scope of the current study, though there are on-going efforts to address the complex relationship between English spelling, pronunciation, and the presence of different sources of natural noise.

NLP models' lack of robustness to noise limits their usage since user-generated inputs can be noisy.
Yet, sacrificing performance on clean inputs to increase robustness is also unacceptable as user-generated inputs can also be clean.
We propose a input-sanitizing model named \MNAME~to help deployed NLP models process clean and noisy user-generated text.
By combining finite-context aggregation with restricted output space, \MNAME~largely preserves the semantic content of the input while imparting reasonable robustness to NLP models.
Thus, \MNAME~can be applied to tasks other than classification, where task-completion requires precise semantic content such as named entity recognition or summarization.
Experimental results show \MNAME~outperforming competitive baselines on six classification tasks and one sequence labeling task under various noise conditions.
On-going work focus on extending \MNAME~to other noisy inputs such as social media text~\cite{derczynski2017results} or conversation transcripts~\cite{kaplan2020may,nguyen2021improving,fu2022effective}.

\section*{Limitations}
In this work, we focused on improving NLP models resistance to noisy input due to realistic adversarial misspellings.
However, natural noise include other types beyond misspellings.
For example, natural noise includes the use of emoticons (e.g. <3), abbreviations (e.g.~`lol'), wordplay using phonetic spelling, mixed casing (capitalization) (e.g.~`so COOOOL'), LEET words~\cite{perea2008r34d1ng} (e.g.~`b4', `R34D1NG'\ldots).
Tackling natural noise would require integrating more explicit visual, phonemic and linguistic knowledge into modeling~\cite{belinkov2018synthetic}.
Besides, the clusters used by \MNAME~are hand-crafted.
Learning the clusters from data may allow models to adapt more quickly to noisy data.

\section*{Acknowledgements}
We thank Gia H. Ngo and the anonymous reviewers for their helpful and constructive feedback.

\bibliography{nlp}
\bibliographystyle{acl_natbib}

\end{document}